\documentclass[letterpaper,10pt,conference]{ieeeconf}  
\IEEEoverridecommandlockouts            
\usepackage{cite}  
\usepackage{epsf}  
\usepackage{graphicx} 
\usepackage{color}
\usepackage{amssymb} 
\usepackage{wrapfig} 
\usepackage{mathtools} 
\usepackage{subfigure,bbm} 
\usepackage{amsmath} 
\usepackage{color} 
\usepackage{subcaption}  
\usepackage[font=footnotesize]{caption} 
\usepackage[colorlinks,linkcolor=blue,anchorcolor=blue,citecolor=blue]{hyperref} 
\usepackage{algorithmic} 
\usepackage{array} 
\usepackage{cancel} 
\usepackage{graphicx} 
\usepackage{color}
\usepackage{amssymb} 
\usepackage{wrapfig} 
\usepackage{mathtools} 
\usepackage{color}  
\usepackage{subfigure,bbm} 
\usepackage{amsmath,amsfonts,bm}
\usepackage{xspace}  
\usepackage{algorithm}   
\usepackage{booktabs}  
\usepackage{multirow}  
\newcommand{\argmax}{\mathop{\mathrm{arg\,max}}}

{\begin{list}{$\bullet$}{%
    \setlength{\topsep}{0in}
    \setlength{\partopsep}{0in}
    \setlength{\itemsep}{0in}
    \setlength{\parsep}{0in}
    \setlength{\leftmargin}{3.5em}
    \setlength{\rightmargin}{0in}
    \setlength{\itemindent}{0in}
}
}%
{\end{list}}

\def\eqref#1{equation~\ref{#1}}
\def\1{\bm{1}}

\DeclareMathAlphabet{\mathsfit}{\encodingdefault}{\sfdefault}{m}{sl}
\SetMathAlphabet{\mathsfit}{bold}{\encodingdefault}{\sfdefault}{bx}{n}

\def\btau{\boldsymbol{\tau}} 
 
\def\btheta{\boldsymbol{\theta}}

\def\bpi{\boldsymbol{\pi}}

\def\md{\mathcal{D}}

\def\ba{\boldsymbol{a}}

\def\bs{\boldsymbol{s}}  
\def\bz{\boldsymbol{z}}

\def\bu{\boldsymbol{u}}  
  
\def\bo{\boldsymbol{o}}  
\def\bc{\boldsymbol{c}}  
\def\bm{\boldsymbol{m}}

\def\btau{\boldsymbol{\tau}} 
 
\def\btheta{\boldsymbol{\theta}}

\def\bpi{\boldsymbol{\pi}}

\newcommand{\ea}[1]{\begin{equation}\begin{aligned}\centering \small #1 \end{aligned}\end{equation}}

\begin{document}   
\title{
\LARGE\bf
Learning a Speed-adaptive Hip Exoskeleton Control Policy Via Sim-to-real Reinforcement Learning  
}  

\author{
Bin~Li$^{*}$, 
Zhimin~Hou$^{*\dag}$, 
Jiacheng~Hou$^{*}$, Zenian~Liang, Tong~Wu, Teng~Ma, and 
Chenglong Fu 
\thanks{
Bin~Li, Zhimin~Hou, Jiacheng~Hou, Zenian~Liang, and Tong~Wu are with Division of Industrial Data Science~(DIDS), School of Data Science, Lingnan University, Hong Kong.}  
\thanks{
Teng~Ma and Chenglong~Fu are with Department of Mechanical and Energy Engineering, Southern University of Science and Technology, Shenzhen, 518055, China. 
}  
\thanks{$*$ The authors contribute equally.} 
\thanks{$\dag$ Corresponding author: Zhimin Hou (e-mail: zhiminhou@ln.edu.hk).}  
}  

\maketitle  
\vspace{-5mm}  
\begin{abstract}  
Providing personalized exoskeleton assistance across varying walking speeds remains challenging. Existing online optimization methods are sample-inefficient, requiring extensive human-in-the-loop~(HIL) evaluations to optimize the entire assistive torque profile. Sim-to-real reinforcement learning~(RL) offers a promising alternative but cannot directly account for individual user preferences. We propose a framework integrating sim-to-real RL with online preference learning for personalized exoskeleton assistance. Specifically, assistance timing is learned in simulation by training RL policies with human musculoskeletal models across varying walking speeds. The learned policies are then distilled and deployed on a physical hip exoskeleton using onboard sensory observations. Gaussian-process-based preference learning further personalizes the assistance magnitude through pairwise user comparisons. By decoupling assistance timing learning in simulation from magnitude optimization in real-world experiments, our framework substantially reduces the online optimization space. Human-subject experiments demonstrate efficient identification of personalized assistive torque profiles across varying walking speeds with fewer real-world evaluations. 
\end{abstract}  

\section{Introduction}\label{sec-intro} 
Exoskeletons hold great promise for reducing physical effort during locomotion by providing assistive joint torques, with applications ranging from mobility assistance to human performance augmentation~\cite{luo2024experiment,molinaro2024task}. 
However, their widespread adoption beyond laboratory environments remains limited due to challenges in developing personalized strategies~\cite{lee2024ai,molinaro2024estimating}, as both assistance timing and magnitude need to be personalized. 
In particular, personalized assistance that remains effective across continuously varying walking speeds is still underexplored.
% Human-in-the-loop optimization was explored to optimize parameterized assistive profiles based on human feedback~\cite{slade2024human}. 

First, timing of assistance is critical for aligning with human motion intention during varying walking speeds\cite{slade2024human}. 
Most exoskeleton controllers have been developed following a hierarchical architecture~\cite{baud2021review}, wherein a high-level controller is responsible for recognizing the user's intent. 
For instance, the gait phase needs to be well detected and the assistive torque profile is designed according to gait phase. 
Model-dependent off-the-shelf spline controllers~\cite{medrano2023real} rely on time-based gait phase estimation~(TBE) method to generate the assistive torque. 
In addition, onboard sensing data, like IMU-based kinematics, was utilized to detect the gait phase variable~\cite{kang2021real}. 
Deep-learning-based model-free methods, including deep neural networks~(DNNs), convolutional neural networks~(CNNs)~\cite{qian2022predictive}, long short-term memory networks~(LSTMs)~\cite{lee2021gaitphase}, and temporal convolutional networks~(TCNs)~\cite{deep_learning_2024}, have provided promising solutions for estimating gait phase under different locomotion conditions.
Additionally, the assistive torque profile was derived from estimated biological torque without relying on gait phase estimation~\cite{molinaro2024task},\cite{molinaro2024estimating}. 
Sim-to-real reinforcement learning~(RL) aims to directly learn the exoskeleton control policy mapping from state to motor commands in simulation~\cite{luo2024experiment,barati2026end}. 
However, most existing sim-to-real RL methods have focused only on specific locomotion mode or walking speed. 

%%%%% human-in-the-loop optimization 
Second, human-in-the-loop optimization is usually applied to optimize personalized assistive torque~\cite{ding2018human} according to human biomechanical or kinematic feedback. 
Metabolic cost has commonly been used as the objective for optimizing parameterized assistive torque~\cite{slade2022personalizing}, defined by timing and magnitude values. 
Probabilistic methods such as Bayesian optimization can efficiently search the assistive parameter space from a limited number of human–exoskeleton evaluations.
Additionally, a policy iteration method, using least squares regression, was integrated with a finite state machine impedance controller~(FSM-IC) to derive a personalized assistive strategy~\cite{li2021toward},\cite{zhang2024toward}. 
Meanwhile, human preference-based learning algorithms have been proposed for personalizing exoskeleton assistance based on measured users' preferences~\cite{ingraham2022role,arens2025preference}. 
However, a major limitation of existing human-in-the-loop optimization methods is their time-consuming nature, with performance highly dependent on the parameterization of strategy. 

\begin{figure*}[!t]  
\setlength{\abovecaptionskip}{-0.01cm}  
\centering  
{\includegraphics[width=1.0\linewidth]{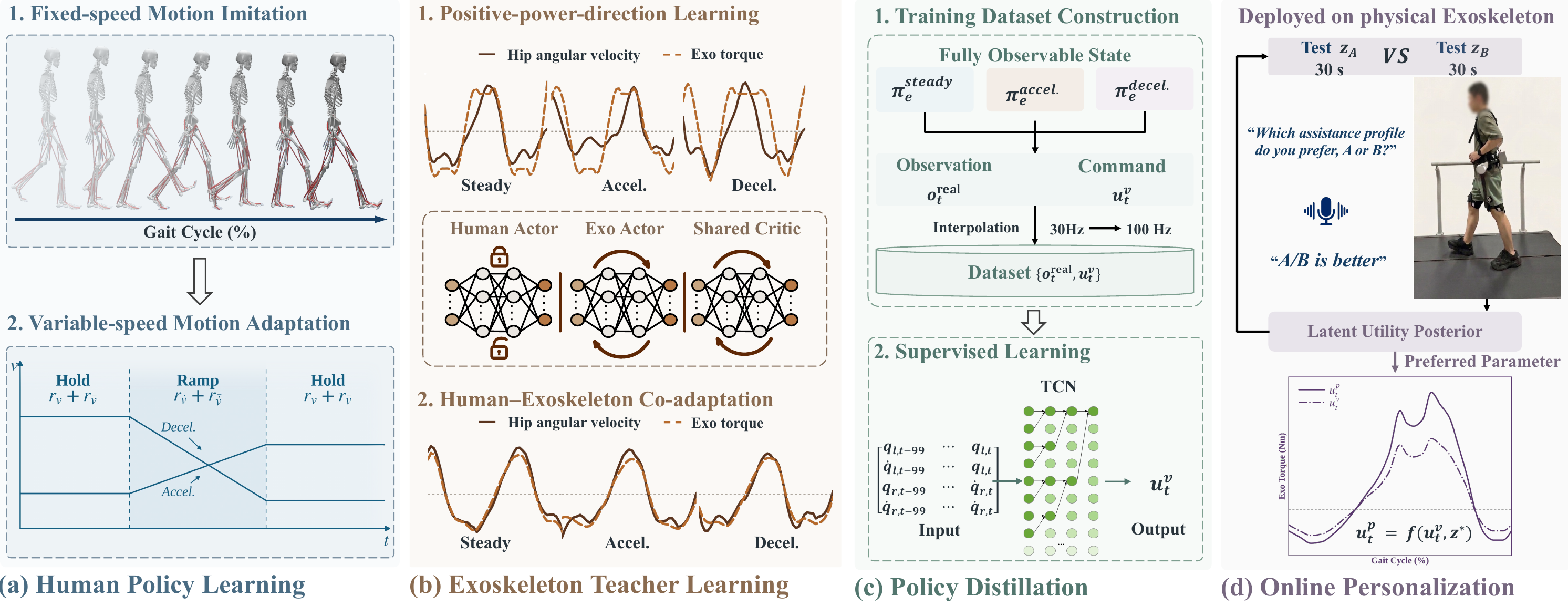}}   
\caption{
Overview of the proposed framework. (a) Learning a human control policy across varying walking speeds. (b) Learning hip exoskeleton teacher policies through interaction with the human musculoskeletal model. (c) Distilling the fully observed teacher policies into a deployable TCN-based student policy using onboard sensory observations. (d) Online preference-based personalization of assistance magnitude after deployment on the physical exoskeleton.
}
\label{fig:overview-framwork} 
\vspace{-0.6cm}  
\end{figure*}   
This work develops a speed-adaptive hip exoskeleton control framework that integrates sim-to-real RL with preference-based optimization. 
\textit{We decouple the optimization of the timing and magnitude of assistive torque profiles across varying walking speeds.} 
First, assistance timing is learned entirely in simulation without explicit gait-phase estimation. 
We train a human control policy capable of locomotion at continuously varying walking speeds and subsequently learn an exoskeleton control policy through interaction with the simulated human. 
Second, the learned exoskeleton policy is distilled into a TCN using onboard sensory observations and deployed on a physical hip exoskeleton. 
Preference-based optimization is then performed for each user to personalize the magnitude of the assistive torque profile. 
The proposed method is validated with six able-bodied participants during walking at varying speeds. 

\section{Related Works} 
\subsection{Sim-to-real RL Techniques for Hip Exoskeleton Control} 
% Sim-to-real RL enables exoskeleton control policies to be trained in simulation, reducing the reliance on extensive real-world experiments~\cite{luo2024experiment}. 
Several musculoskeletal models and simulation platforms have been developed to reproduce human biomechanics and locomotion~\cite{zuo2024self,tan2025myoassist}. 
Predictive neuromusculoskeletal simulation model was employed to train exoskeleton policies for assistive control~\cite{park2026learning}. 
Alternatively, deep RL-based human control policies have been developed to actuate musculoskeletal models and reproduce desired human motions~\cite{lee2019scalable,weng2021natural,luo2024experiment}, providing interactive human models for training exoskeleton policies. 
Based on such simulations, RL-based exoskeleton policies have been optimized to reduce human muscular effort~\cite{barati2026end} and successfully transferred to physical exoskeletons, with their assistance effectiveness validated on human participants~\cite{park2026learning,barati2026end}. 
\textit{However, existing RL-based human control policies typically rely on predefined reference motions to reproduce human-like gait, limiting their ability to represent natural locomotion under continuously varying walking speeds.} 
\subsection{Personalized Exoskeleton Control Optimization} 
% Human-in-the-loop (HIL) learning has been widely investigated for online adaptation and optimization of personalized assistance in lower-limb wearable robots~\cite{zhang2017human,ding2018human}. 
The effectiveness of HIL optimization largely depends on the learning objective and policy parameterization~\cite{zhang2017human,ding2018human,slade2024human}. 
The objective is typically defined using measurable biomechanical or kinematic feedback, while policy parameterization determines the adaptable degrees of freedom of the assistive strategy. 
For example, assistive torque profiles were parameterized by a compact set of timing and magnitude variables. 
CMA-ES and Bayesian optimization have been employed to optimize such parameters for reducing metabolic cost~\cite{zhang2017human,park2025human}. 
Gradient-based methods, including policy iteration, have been developed to improve gait symmetry or track performance~\cite{li2021toward,zhang2024toward}.  
However, optimizing high-dimensional assistive policies typically requires extensive human–robot interaction trials, particularly for sample-based optimization methods. 
Preference-based optimization has recently been explored to reduce the reliance on explicitly defined quantitative objectives~\cite{tucker2020human,ramella2025rapid,lee2023user}. 
Instead of requiring numerical performance feedback, these methods infer user preferences from pairwise comparisons between candidate assistance strategies, providing an intuitive interface for personalized optimization. 
\textit{Nevertheless, existing HIL-based methodes primarily optimize assistive profiles for a specific locomotion condition, while personalized assistance across varying walking speeds remains underexplored.}  
% \vspace{-0.6cm}  
\section{Overview Framework}\label{sec-problem}  
This work aims to integrate offline learning of \textbf{assistance timing} via sim-to-real RL and preference-based online optimization of \textbf{assistance magnitude}. 
The overview framework is shown in Fig.~\ref{fig:overview-framwork}.  
A MuJoCo-based simulation platform was developed for modeling dynamic human–exoskeleton interaction~\cite{tan2025myoassist}. 
A human musculoskeletal model, consisting of a skeleton model with $n_h=53$ degrees of freedom~(DOFs) and $n_m=22$ muscle tendon units, was employed for lower-limb locomotion reproduction. 
The human model was constrained to move in the sagittal plane and was driven by muscle activation $\ba \in \mathbb{R}^{n_m}$. 
The joint kinematics and biological torque $\btau_h \in \mathbb{R}^{n_h}$ can be measured at each time step. A virtual hip exoskeleton driven by $\btau_e \in \mathbb{R}^{n_e}$ for flexion and extension assistance with $n_e =2$ DOFs was employed as an example to demonstrate the effectiveness of the proposed method. 
% $n_e \ll n_h$ is the dimension of actuation joints for the exoskeleton. 
The implementation of our method will be divided into four stages. 
First, a human control policy $\bpi_h$ that can drive the human musculoskeletal model to reproduce locomotion at varying walking speeds was learned via in-context reinforcement learning. 
Second, an exoskeleton control policy $\bpi_e^{\textrm{sim}}$ was learned to reduce muscle effort across varying walking speed. 
Third, a RL-based exoskeleton policy was distilled into $\bpi_e^{\textrm{real}}$ and deployed on physical exoskeleton. 
Finally, preference-based online optimization method was developed to optimize the magnitude-related parameters. 

\section{Method}\label{sec-method}  
\subsection{Human Control Policy with Variable Walking Speed}\label{subsec-interactive-control} 
% A human control policy is parameterized as $\bpi_h(\ba|\bs,\bc;\btheta)$. $\bc$ is used to represent the task context, walking speed. 
% The action space of the human control policy is defined by the muscle activation
% $\ba \in [0,1]^{22}$. 
% In context reinforcement learning, similar to \cite{chiu2025learning,tang2024humanmimic}, . 
% The human control policy was trained via two separate phases, represented as $\bpi_h^{p_1}$ and $\bpi_h^{p_2}$. 
% At the first phase, learns a stable reference-guided gait at a fixed walking speed. 
% At the second phase adapts the learned locomotion policy to variable-speed walking through velocity- and acceleration-conditioned control. 
% $\bpi_h^{p_1}$ learns a stable fixed-speed gait at by tracking reference joint angles and velocities. 
% $v$ is the forward pelvis
% velocity. 
% The reference-motion imitation terms used in the first training phase are disabled In \textit{Phase} 2. 
A context-conditioned human control policy is parameterized as $\bpi_h(\ba|\bs,\bc;\btheta_h)$, where $\bc$ represents the task context characterizing the desired locomotion condition. 
The policy outputs muscle activations $\ba$ to drive the human musculoskeletal model. 
Similar to context-conditioned RL methods~\cite{chiu2025learning,tang2024humanmimic}, the task context is incorporated into the policy observation to enable a unified policy to generalize across multiple locomotion tasks. 
The human control policy is trained via two phases, resulting in $\bpi_h^{p_1}$ and $\bpi_h^{p_2}$, respectively. 
In \textit{Phase} 1, $\bpi_h^{p_1}$ learns a stable fixed-speed gait by tracking reference joint angles and velocities, where $\bc$ is the commanded forward pelvis walking velocity. 
In \textit{Phase} 2, the policy $\bpi_h^{p_2}$ is trained for variable-speed locomotion given the context of the target walking velocity $v^{\textrm{goal}}$ and acceleration $\dot{v}^{\textrm{cmd}}$. 
The reference-motion imitation terms are removed during \textit{Phase 2}, allowing $\bpi_h^{p_2}$ to generate adaptive locomotion behaviors across different velocity conditions without explicit motion tracking. 
\subsubsection{Observation Space} 
In \textit{Phase} 1, the observation $\mathbf{o}_h^{p_1}$ is defined as
\begin{equation} 
\mathbf{o}_h^{p_1}
=
[\mathbf{q},\dot{\mathbf{q}},\ba,\mathbf{f},v^{\mathrm{cmd}}] 
\in\mathbb{R}^{44}. 
\end{equation}
where $\mathbf{q}$ and $\dot{\mathbf{q}}$ denote the selected joint angles and velocities. $\mathbf{f}$ is the bilateral foot-contact force. 
$v^{\mathrm{cmd}}$ is the commanded walking velocity. 

In \textit{Phase} 2, the observation is defined as: 
\begin{equation}
\mathbf{o}_h^{p_2}
=
[
\mathbf{o}_h^{p_1},\,
\dot{v}^{\mathrm{cmd}},\,
v^{\mathrm{goal}}
]
\in \mathbb{R}^{46}.
\label{eq:h2_obs}
\end{equation}
where $\dot{v}^{\mathrm{cmd}}$ is the commanded acceleration and $v^{\mathrm{goal}}$ is the target velocity. 
\subsubsection{Reward Function} 
In \textit{Phase} 1, 
$r_h^{p_1}$ is defined as
\begin{equation}
\begin{aligned}
r_h^{p_1} ={}&
w_q r_q + w_{\dot q} r_{\dot q} + w_v r_v + w_m r_m + w_{\Delta m} r_{\Delta m}\\
&
+ w_{\textrm{constraint}} r_{\textrm{constraint}} + w_{\textrm{foot}} r_{\textrm{foot}}. 
\end{aligned}
\label{eq:h1_reward}
\end{equation} 
where $r_q$ and $r_{\dot q}$ are terms to encourage the reference motion imitation, defined as: 
\begin{equation}
\begin{aligned} 
r_q &=
\sum_{i \in n_h} c_i^q \Delta t
\exp[-8(q_i-q_i^{\mathrm{ref}})^2], \\
r_{\dot q} &=
\sum_{i \in n_h} c_i^{\dot q}\Delta t
\exp\left[
-8\left(
\dot q_i
-\rho \dot q_i^{\mathrm{ref}}
\right)^2
\right],
\end{aligned}
\label{eq:h1_imitation_rewards}
\end{equation}
where $i \in n_h$ is the index of
tracked joint, $c_i^q$ and $c_i^{\dot q}$ are the corresponding
joint-specific weights, and $\rho =
v^{\mathrm{cmd}}/\dot q_{\mathrm{pelvis}}^{\mathrm{ref}}$ scales the
reference joint velocities according to the commanded walking speed. $\Delta t$ is the control step. 
$r_v$ is defined to keep the current walking speed as follows: 
\ea{
r_v = \Delta t \exp[-5(v-v^{\mathrm{cmd}})^2], 
}
where $v$ is the actual forward walking velocity. 
The remaining auxiliary reward terms regularize muscle effort, activation smoothness, joint constraints, and ground-contact forces: 
\begin{equation}
\begin{aligned} 
r_m &=
-\Delta t/n_m\sum\nolimits_{j=1}^{n_m} a_{j,t}, \\
r_{\Delta m} &=
\Delta t/n_m\sum\nolimits_{j=1}^{n_m}
\exp[-4(a_{j,t}-a_{j,t-1})^2], \\
r_{\mathrm{constraint}} &=
-\Delta t\,
\max_{k}
\left(
|f_{k}^{\mathrm{joint}}|/{Mg}
\right), \\
r_{\mathrm{foot}} &=
-\Delta t\,
\max
\left(
0,\,
\frac{|f_r|+|f_l|}{Mg}-1.2
\right).
\end{aligned}
\label{eq:h1_regularization_rewards}
\end{equation} 
where 
$a_{j,t}$ and $a_{j,t-1}$ denote the activation of muscle $j$ at the current and previous control steps, respectively. 
$r_{\mathrm{constraint}}$ penalizes the maximum joint constraint force $f_k^{\mathrm{joint}}$ among the monitored joints, normalized by the body weight $Mg$. $r_{\mathrm{foot}}$ penalizes the total bilateral foot contact force exceeding $1.2$ times the body weight. 

In \textit{Phase} 2, $\bpi_h^{p_2}$ was trained to reproduce the 
variable-speed locomotion by tracking the commanded velocity during hold phases and the commanded acceleration during speed transitions. 
$v_t^{\mathrm{cmd}}$ is defined as: 
\begin{equation}
v_t^{\mathrm{cmd}}
=
\begin{cases}
{v}^{\mathrm{init}}, & t<t_s,
\\
{v}^{\mathrm{init}}+\dot{v}^{\mathrm{cmd}}(t-t_s),
& t_s\leq t<t_s+T_r,
\\
v^{\mathrm{goal}},
& t\geq t_s+T_r .
\end{cases}
\label{eq:variable_speed_command}
\end{equation}
where $t_s$ denotes the onset of the speed transition, and
$T_r = |v^{\mathrm{goal}}-{v}^{\mathrm{init}}|/|\dot{v}^{\mathrm{cmd}}|$ is constrained to 
$1$--$5~\mathrm{s}$. 
The initial and target speeds are selected from a
predefined discrete set of representative speeds spanning
$0.5$--$1.7~\mathrm{m/s}$. 
$\dot{v}^{\mathrm{cmd}}$ is similarly selected
from a finite set of acceleration/deceleration levels within
$[-0.5,\,0.5]~\mathrm{m/s^2}$. 
$r_h^{p_2}$ is defined as 
\begin{equation}
\begin{aligned}
r_h^{p_2} ={}&
\chi^{\mathrm{hold}}
\left(w_v r_v + w_{\bar v} r_{\bar v}\right)
+\chi^{\mathrm{ramp}}
\left(w_{\dot v} r_{\overline{\dot v}} + w_{\dot v} r_{\overline{\dot v} }\right) \\
&+
w_m r_m .
\end{aligned}
\label{eq:h2_reward}
\end{equation}
where $\chi^{\mathrm{hold}}=1$ is set for the initial and final hold phases. 
$\chi^{\mathrm{ramp}}=1$ is set the speed-transition phase. 
The step-level velocity tracking
$r_{\bar v}$ and local and average acceleration tracking
$r_{\dot v}$ and $r_{\overline{\dot v}}$ are introduced in \textit{Phase} 2, as follows: 
\begin{equation} 
\begin{aligned}
r_{\bar v}
&=
-\frac{\Delta t}{L_{\mathrm{leg}}}
\left|
\sum_{\ell \in\mathcal{S}_n}
\Delta t\,(v_\ell-v_\ell^{\mathrm{cmd}})
\right|,\\
r_{\dot v}
&=
\Delta t
\exp[-8\left(\dot v^{\mathrm{est}}-\dot v^{\mathrm{cmd}}\right)^2],\\
r_{\overline{\dot v}}
&=
\Delta t
\exp\left[
-8\left(
\overline{\dot v}-\dot v^{\mathrm{cmd}}
\right)^2
\right].
\end{aligned}
\label{eq:h2_reward_terms}
\end{equation}
where $L_{\mathrm{leg}}$ is the leg-length normalization factor. 
$S_n$ denotes the samples within the $n$-th gait step.  
$\dot v^{\mathrm{est}}$ is estimated from a $0.3$s velocity window. $\overline{\dot v}$ is the average acceleration over the corresponding gait-step interval.
% \begin{figure}[!t]  
% \setlength{\abovecaptionskip}{-0.05cm}  
% \centering  
% {\includegraphics[width=1.0\linewidth]{figures/AAN/figure_2_b.pdf}}   
% \caption{
% Illustration of patient motion preferences encoding, therapist-informed via-point extraction, and reference trajectory deformation in the proposed framework.~(a)~Gray lines are patient's actual motion trajectories. The green line indicates the estimated mean motion trajectory according to Section~\ref{subsec-motion-preference-encoding}. Red scatter points indicate therapist-informed via-points inferred from the therapist-informed corrective force and the desired motion. The blue line indicates the generated reference trajectory according to Section~\ref{subsec-trajectory-deformation}.~(b)~Gray lines represent actual motion trajectories reproduced by therapist-side robot, while therapist applies the effective corrective force as the red arrows to inform the via-points. 
% }  
% \label{Fig2:trajectory-deformation} 
% \vspace{-0.6cm}  
% \end{figure}   

\subsection{Exoskeleton Control Policy Learning}\label{subsec-exo-policy-learning}  
The learned human control policy $\bpi_h$ can generate locomotion across varying walking speeds. 
Afterwards, we train an exoskeleton teacher policy $\bpi_e(\ba^e|\bs^e;\btheta_e)$ via continuous action RL methods to reduce human muscle effort, as illustrated in Fig.~\ref{fig:overview-framwork}(b). 
A virtual bilateral hip exoskeleton with a maximum actuator torque $\btau_e^{\mathrm{MAX}}$ is coupled with the musculoskeletal model and controlled by $\bpi_e$. 
The exoskeleton policy is trained in two phases, resulting in $\bpi_e^{p_1}$ and $\bpi_e^{p_2}$. 
The policy outputs normalized bilateral hip commands $[u_r, u_l]\in[-1,1]^2$, which are scaled by $\btau_e^{\mathrm{MAX}}$ to generate the assistive torques. 
The human and exoskeleton policies share a common critic network during training. In \textit{Phase} 1, the pretrained human control policy $\bpi_h$ is frozen, while $\bpi_e^{p_1}$ is optimized to provide assistance that reduces muscle effort while adapting to the learned human locomotion. 
In \textit{Phase} 2, the human control policy and $\bpi_e^{p_2}$ are jointly optimized, enabling human–exoskeleton co-adaptation to further refine the assistance profile. 
\subsubsection{Observation Space}  
% Together with the human muscle commands, the joint
% action is
% \begin{equation}  
% \mathbf{u_e} 
% =[\mathbf{a},\,u_r,\,u_l] 
% \in \mathbb{R}^{24}.
% \label{eq:human_exo_action}
% \end{equation} 
In \textit{Phase} 1, the observation for exoskeleton control teacher policy training is defined as: 
\begin{equation} 
\mathbf{o}_{e}^{p_1} 
=[\mathbf{o}_{h}^{p_2}, \mathcal{U}_{e}]
\in \mathbb{R}^{52}. 
\label{eq:e1_obs}
\end{equation} 
where $\mathcal{U}_{e} \in \mathbb{R}^{6}$ is a three-step history of the bilateral normalized exoskeleton commands. $\mathbf{o}_{h}^{p_2}$ is the human state used in \textit{Phase} 2 of human control policy learning. 

In \textit{Phase} 2, a fully observed state is defined as: 
\begin{equation}  
\mathbf{o}_{e}^{p_2}=[\mathbf{o}_{e}^{p_1},\, u_{r,t-1}, u_{l,t-1}] \in \mathbb{R}^{54}. 
\label{eq:e2_obs}  
\end{equation}  
where $[u_{r,t-1}, u_{l,t-1}] \in \mathbb{R}^{2}$ denotes the bilateral torque commands applied at the previous control step. 

\subsubsection{Reward Function} 
In \textit{Phase} 1, $\bpi_e^{p_1}$ aims to learn an assistance pattern by favoring exoskeleton torques that generate mechanical power in the desired direction. Separate exoskeleton control policies are trained for steady walking, acceleration, and deceleration, with the assistive torque limited to $6~\mathrm{Nm}$. $r_e^{p_1}$ is defined as 
\begin{equation} 
\begin{aligned} 
r_e^{p_1}
={}& r_h^{p_2}
+ w_P r_P
+ w_{\Delta m} r_{\Delta m} \\
&+ w_{\mathrm{constraint}} r_{\mathrm{constraint}}
+ w_{\mathrm{foot}} r_{\mathrm{foot}} 
\end{aligned}
\end{equation} 
where $r_h^{p_2}$ with $\chi_t^{\mathrm{hold}}=1$ is added to encourage the steady walking. 
$r_P$ encourages the desired power-direction and
penalizes assistance in the opposite direction, as follows: 
\begin{equation}
r_P
=
\Delta t
\sum\nolimits_{s\in\{l,r\}}
\alpha_P u_{s}^{2}
\operatorname{sign}
\!\left({u}_s
\dot{q}_s^{\mathrm{hip}}
\right).
\label{eq:e1_power_reward}
\end{equation}
where $u_s$ denotes the normalized exoskeleton torque command
on side $s$. 
$\dot{q}_s^{\mathrm{hip}}$ is the hip angular velocity on side $s$. 
For the steady teacher, both $r_{\dot v}$ and
$r_{\bar{\dot v}}$ remain inactive because no speed transition is defined.

In \textit{Phase} 2, the human and exoskeleton actors are jointly optimized to refine the assistance pattern. The assistive torque limit is set to $12~\mathrm{Nm}$. 
$r_e^{p_2}$ is defined as 
\begin{equation} 
\begin{aligned} 
r_e^{p_2} ={}&
r_h^{p_2}
+\chi^{\mathrm{st}} w_{\dot v}^{\mathrm{st}} r_{\dot v}
+w_{\mathrm{hip}} r_{\mathrm{hip}}+
w_A r_A  +w_{\Delta u} r_{\Delta u}\\
&
+ w_{\Delta m} r_{\Delta m}
+ w_{\mathrm{constraint}} r_{\mathrm{constraint}}
+ w_{\mathrm{foot}} r_{\mathrm{foot}}. 
\end{aligned}
\label{eq:e2_reward}
\end{equation}
where $r_P$ is disabled by setting $\omega_P = 0$. 
$\chi^{\mathrm{st}}=\mathbb{I}(g=\mathrm{steady})$ is the
steady-teacher indicator. 

Here, $r_{\mathrm{hip}}$ penalizes hip-related muscle activation,
$r_A$ promotes effective assistance while suppressing excessive torque, and 
$r_{\Delta u}$ regularizes temporal variations in the commanded torque, as follows: 
\begin{equation} 
\begin{aligned}
r_{\mathrm{hip}}
&=
-\Delta t/N_{\mathrm{hip}}
\sum\nolimits_{m=0}^{N_{\mathrm{hip}}} a_m^{2},
\\
r_A
&=
\Delta t
\sum\nolimits_{s\in\{l,r\}}
\left[
\alpha u_s{\hat{\dot{q}}_s^{\mathrm{hip}}}
-\beta u_s^{2}
-\lambda\max(0,| u_s|-\delta)^2
\right],
\\
r_{\Delta u}
&=
-\Delta t
\sum\nolimits_{s\in\{l,r\}}
(u_{s,t}-u_{s,t-1})^2 .
\end{aligned}
\label{eq:e2_reward_terms}
\end{equation}
where $\mathcal{M}_{\mathrm{hip}}$ denotes the hip-related muscle set. 
$N_{\mathrm{hip}}=|\mathcal{M}_{\mathrm{hip}}|$ is the number of hip-related muscles. 
$\hat{\dot{q}}_s^{\mathrm{hip}}=\operatorname{clip}
(\dot{q}_{\mathrm{hip}}/\dot{q}_{\mathrm{scale}},-1,1)$ is the normalized hip
angular velocity by $\dot{q}_{\mathrm{scale}}=2.0~\mathrm{rad/s}$. 

% \begin{figure}[htbp]  
% \setlength{\abovecaptionskip}{-0.02cm}  
% \centering  
% {\includegraphics[width=1.0\linewidth]{figures/fig2.pdf}}   
% \caption{
% Speed-adaptive behavior of the learned human motion policy. Hip, knee, and ankle joint kinematics and moments across walking speeds from 0.5 to 1.7 m/s, with orange curves denoting the reference human motion. 
% }
% \label{fig:human-performance-multiple-speed}  
% \vspace{-0.6cm}  
% \end{figure}   
\begin{figure*}[!t]
\centering
\setlength{\abovecaptionskip}{-0.2cm}
% -------------------- (a) --------------------
\begin{minipage}[t]{0.71\linewidth}
    \centering
    \includegraphics[
        height=5.25cm,
        keepaspectratio
    ]
    {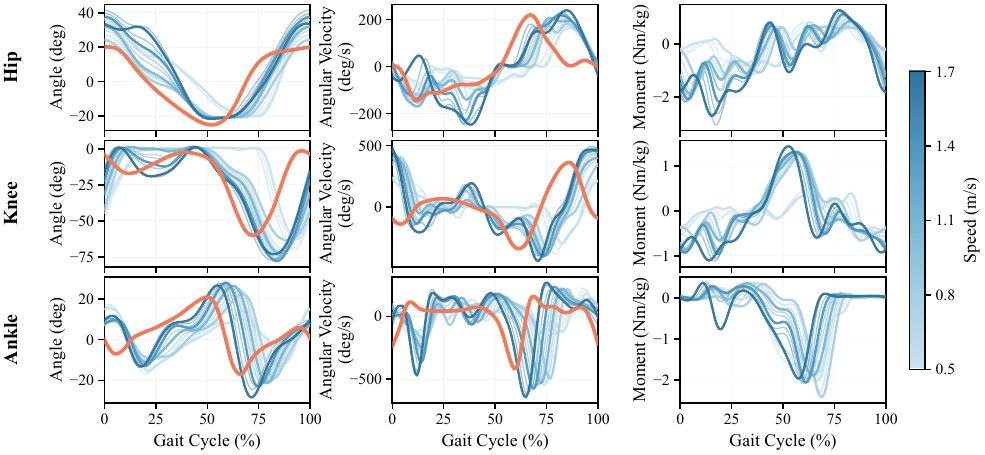}
    
    \vspace{-0.18cm}
    {\small (a)}
    \label{fig:human-performance-multiple-speed}
\end{minipage}
% \hspace{0.015\linewidth}
% -------------------- (b) --------------------
\begin{minipage}[t]{0.26\linewidth}
    \centering
    \includegraphics[
        height=5.25cm,
        keepaspectratio
    ]{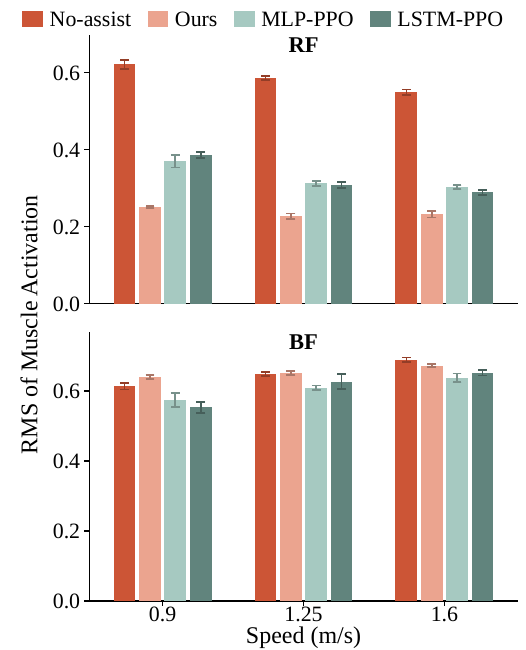}
    \par
    \vspace{-0.18cm} 
    \hspace*{0.35cm}{\small (b)}
    \label{fig:muscle_activation}
\end{minipage}
\vspace{0.1cm}
\caption{
Simulation results of human motion reproduction and muscle activation.
(a) Speed-adaptive behavior of the learned human motion policy. Hip, knee, and ankle joint kinematics and moments across walking speeds from ${0.5}$ to $1.7~\mathrm{m/s}$, with orange curves denoting the reference motion;
(b) Comparison of RMS muscle activation across assistance methods at ${0.9}$,${1.25}$, and $1.6~\mathrm{m/s}$.
}
\label{Fig6:comparison-with-baselines} 
\vspace{-0.5cm}
\end{figure*}   
% \begin{figure*}[!t]  
% % \setlength{\abovecaptionskip}{-0.05cm} 
% \centering  
% {\includegraphics[width=1.0\linewidth]{figures/fig3.png}}     
% \caption{
% Overview of learning our shape-adaptive AAN rehabilitation policy based on two isomorphic robots interacting with patient and therapist through the end-effector, separately. 
% }
% \label{fig:exo-performance-variable-speed}  
% \vspace{-0.4cm}    
% \end{figure*}  
\begin{figure*}[!t]
\centering 
\setlength{\abovecaptionskip}{-0.025cm}  
\includegraphics[width=\linewidth]{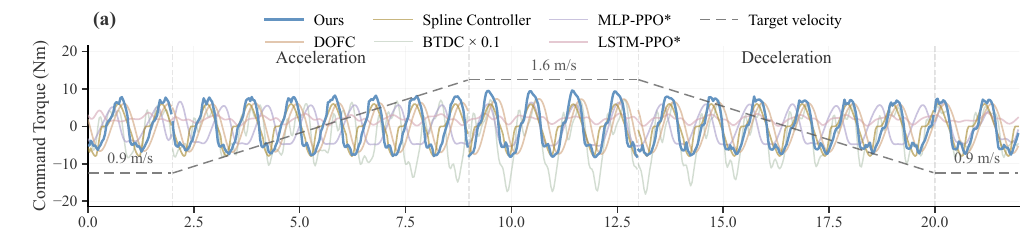} 
\includegraphics[width=\linewidth]{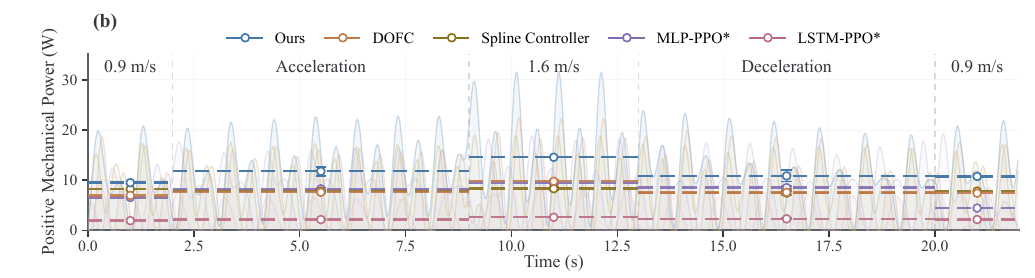} 
\caption{
Comparison of assistance performance during variable-speed walking~(acceleration from $0.9~\mathrm{m/s}$ to $1.6~\mathrm{m/s}$ and deceleration from $1.6~\mathrm{m/s}$ to $0.9~\mathrm{m/s}$).(a) Commanded assistive torque over the speed-transition sequence. (b) Corresponding positive mechanical power and gait-level mean positive power (MPP).
The MLP-PPO and LSTM-PPO policies are trained only at a fixed walking speed of $1.25~\mathrm{m/s}$ and are directly applied to the acceleration and deceleration stages for comparison. 
}
\label{fig:exo-performance-variable-speed}
\vspace{-0.7cm}
\end{figure*}  
\subsection{Exoskeleton Control Policy Distillation} 
After training the three fully observed teacher policies for different locomotion conditions, we distill their behaviors into a student policy $\bpi_e^{\textrm{real}}(\bu_t^v|\bo_t^{\textrm{\textrm{real}}};\btheta_e^{\textrm{real}})$ that relies only on onboard sensory observations.
A temporal convolutional network~(TCN) is adopted as the student policy to handle the partially observed onboard state $\bo_t^{\textrm{real}}$. 
Training data pairs $(\bo_t^{\textrm{real}}, \bu_t^v)$ are generated by rolling out each teacher policy given the specific curriculum scheme. 
The observation–action pairs are interpolated from 30 Hz to 100 Hz to construct the training dataset $\md_{\textbf{TCN}}$. 
As shown in Fig.~\ref{fig:overview-framwork}(c), parameters $\btheta_e^{\textrm{real}}$ were updated via supervised learning scheme.  
\subsection{Exoskeleton Control Policy Online Tuning}\label{subsec-trajectory-deformation} 
The distilled $\bpi_e^{\textrm{real}}(\bu_t^v|\bo_t^{\textrm{real}}; \btheta_e^{\textrm{real}})$ exoskeleton control policy determines the assistance timing according to the measurable human state. 
As shown in Fig.~\ref{fig:overview-framwork}(d), the observation $\bo_t^{\textrm{real}} =
[q_{l,t}^{\textrm{hip}}, \dot{q}_{l,t}^{\textrm{hip}},
q_{r,t}^{\textrm{hip}}, \dot{q}_{r,t}^{\textrm{hip}}]$ can be measured from IMUs at 100 Hz and low-pass filtered at 5 Hz. 
The personalized assistive torque profile $\bu_t^p$ was parameterized given the output torque $\bu_t^v$ and variable $\bz$, as follows: 
\ea{
\bu_t^p = f(\bu_t^v, \bz) 
} 
where $\bz \in \mathbb{R}^2$ parameterizes the assistance magnitudes for hip extension and flexion, respectively. 
The assistance timing learned in simulation is preserved, while $\bz$ independently adjusts the extension and flexion magnitudes. 
For instance, the torque profile is revised as illustrated in Fig.~\ref{fig:overview-framwork}(d). 

During online optimization, different assistive torque profiles can be generated given specific $\bz \in \boldsymbol{\mathcal{Z}}$. 
A latent utility function is defined as $\mathcal{R}(\bz)$ to evaluate the preference of each user on selected assistive torque profile. 
The objective of online optimization is to obtain the optimized assistive torque profile $\bz^{*} = \argmax_{\bz} \mathcal{R}(\bz)$. 
At each optimization iteration $k \in K$, the pairwise comparisons with active querying were applied to collect user feedback $\boldsymbol{b}_k$ via voice interaction~(see Fig.~\ref{fig:overview-framwork}(d)).  
The dataset $\md_{\textbf{Pre}} = \md_{\textbf{Pre}} + \{(\bz_k, \boldsymbol{b}_k)\}$ will be updated. 
Then, we utilize Gaussian process to estimate the prior of the latent utility function, as follows: 
\ea{
f(\bz) \sim \mathcal{GP}(\mu(\bz), k(\bz, \bz^{\prime})) 
} 
The latent utility posterior can be inferred from pairwise preferences, as follows: 
\ea{ 
p(f \mid \md_{\textbf{Pre}}) 
\propto
p(f) 
\prod\nolimits_{(\bz_k^A \succ \bz_k^B) 
\in \md_{\textbf{Pre}}} 
\Phi\!\left( 
\frac{
f(\bz_k^A)
-
f(\bz_k^B)
}{\sigma_p}  
\right) 
}  
where $\sigma_p$ is the predefined preference noise. The mean $\mu_k(\bz)$ and variance $\sigma_j(\bz)$ of posterior function is obtained using Laplace approximation. 
For the next optimization iteration, the assistive torque profiles generated based on $\bz_{k+1}^A$ and $\bz_{k+1}^B$ will be evaluated. 
The choice $A$ using the best parameter $\bz_{k+1}^{A} = \argmax \mu(\bz)$ can be obtained and the choice $B$ using another parameter can be obtained according to $\bz_{k+1}^{B} = \argmax [\mu_k(\bz) + \beta \sigma_k(\bz)]$.   
\section{Simulation Validation}\label{sec-simulation-validation} 
The simulation was only utilized for evaluating the performance of assistance timing. 
% The implementation of learning an exoskeleton control policy for varying walking speed was summarized in Algorithm~\ref{alg-exo-policy-learning}. 
\subsection{Simulation Protocol}\label{sec-simulation-protocol} 
The control time steps was set to $\Delta t = 0.0333\mathrm{s}$. 
In principle, continuous action RL algorithms, such as TD3 and SAC, can be used to learn both the human and exoskeleton control policies. 
In this work, PPO was adopted for all policy learning. 
The used hyperparameters of PPO are provided in Table~\ref{tab:ppo_parameters}. 
Since the primary objective of this work is to demonstrate the effectiveness of the proposed framework rather than optimize individual hyperparameters, the reward weights were empirically selected and kept fixed throughout the experiments, as summarized in Table~\ref{tab:reward_weights}. 
\vspace{-2mm}
\subsection{Simulation Results}\label{sec-simulation-results} 
\subsubsection{Performance of Learned Human Control Policy}  
The hip, knee, and ankle joint kinematics in the sagittal plane were evaluated across walking speeds ranging from $0.5~\mathrm{m/s}$ to $1.7~\mathrm{m/s}$.
As shown in Fig.~\ref{Fig6:comparison-with-baselines}(a), the learned human control policy reproduces speed-adaptive locomotion over the entire evaluated speed range, despite being trained with reference motion only at $1.25~\mathrm{m/s}$. 
At the reference speed, the pooled RMSEs across the hip, knee, and ankle were \(17.06^\circ\) for joint angles and \(211.93^\circ/\mathrm{s}\) for angular velocities. Gait cycles were segmented using peaks of the left hip angle and time-normalized to \(0\text{--}100\%\) before RMSE calculation. All metrics are only calculated for the left limb. 
\begin{table*}[!t]
\centering
\caption{Commanded Torque and Mechanical Power Across Subjects} 
\label{tab:experimental_results} 
\scriptsize 
\setlength{\tabcolsep}{2.4pt}
\begin{tabular}{c | cccc | cccc | cccc}
\toprule 
\multirow{2}{*}{Subject}
& \multicolumn{4}{c|}{DOFC}
& \multicolumn{4}{c|}{Ours}
& \multicolumn{4}{c}{Online Tuning} \\
\cmidrule(lr){2-5}
\cmidrule(lr){6-9}
\cmidrule(lr){10-13}
&
$\btau_{\mathrm{RMS}}(Nm)$
&
$\btau_{\mathrm{MAX}}(Nm)$ 
&
MPP
&
MNP
&
$\btau_{\mathrm{RMS}}$ 
&
$\btau_{\mathrm{MAX}}$ 
&
MPP
&
MNP
&
$\btau_{\mathrm{RMS}}$
&
$\btau_{\mathrm{MAX}}$ 
&
MPP
&
MNP
\\

\midrule

% Jiacheng
S1
& 2.57 & 4.04 & 4.30 & -0.090
& 3.85 & 5.25 & 5.26 & -0.004
& 6.35 & 9.96 & 8.63 & -0.192 \\

% Jimmy
S2
& 3.69 & 5.37 & 6.60 & -0.022
& 3.09 & 4.52 & 5.06 & -0.012
& 3.23 & 5.36 & 5.49 & -0.093 \\

% Li Bin
S3
& 3.27 & 4.87 & 6.40 & -0.321
& 3.06 & 4.61 & 5.17 & -0.059
& 5.66 & 8.08 & 10.28 & -0.094 \\

% Wu Tong
S4
& 4.12 & 5.83 & 5.79 & -0.089
& 3.54 & 5.18 & 3.55 & -0.011
& 4.66 & 9.00 & 7.59 & -0.123 \\

% Tihan
S5
& 3.24 & 4.64 & 4.83 & -0.020
& 3.90 & 5.74 & 5.64 & -0.040
& 5.69 & 8.94 & 8.77 & -0.005 \\

% Zenian
S6
& 1.79 & 2.60 & 2.89 & -0.147
& 4.08 & 5.49 & 8.86 & -0.023
& 5.15 & 8.66 & 9.50 & -0.010 \\
\midrule
Mean$\pm$SD
& 3.11$\pm$0.83
& 4.56$\pm$1.14
& 5.14$\pm$1.41
& -0.12$\pm$0.11
& 3.59$\pm$0.43
& 5.13$\pm$0.48
& 5.59$\pm$1.76
& -0.03$\pm$0.02
& 5.12$\pm$1.09
& \textbf{8.33}$\pm$\textbf{1.58} 
& \textbf{8.38}$\pm$\textbf{1.68} 
& -0.09$\pm$0.07\\

\bottomrule
\end{tabular}
\begin{flushleft}
\footnotesize
\end{flushleft}
\vspace{-0.6cm}
\end{table*} 
\begin{figure*}[!t] 
\setlength{\abovecaptionskip}{-0.25cm}   
\centering  
\subfigure[]{\includegraphics[width=0.71\linewidth]{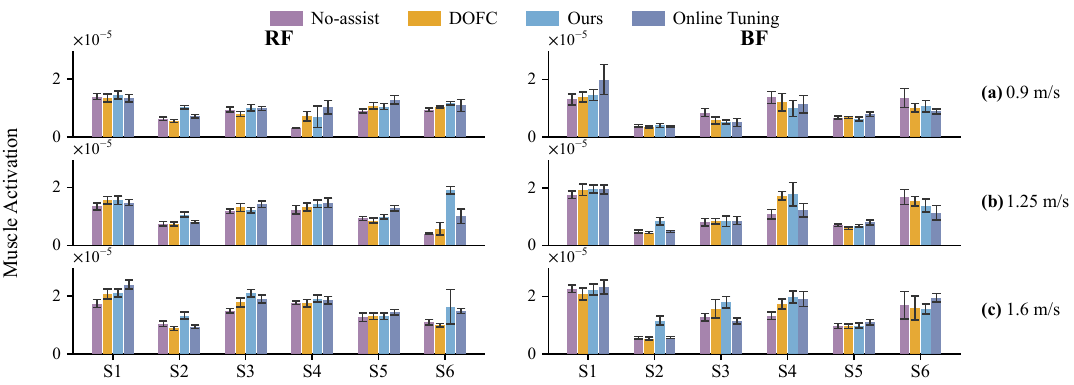}    
\label{Fig6:-emg-all}}    
\subfigure[]{\includegraphics[width=0.23\linewidth]{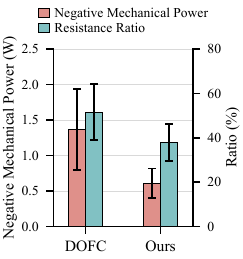}   
\label{Fig6:-power-free-walking}}   
\caption{ 
Experimental results of muscle activation and resistive assistance.~(a)~Comparison of muscle activation across three assistance methods at three walking speeds.~(b)~Resistance ratio, defined as the fraction of samples exhibiting negative assistive mechanical power within 100~ms after hip angular-velocity zero crossings. 
} 
\label{Fig6:emg-comparison-with-baselines}   
\vspace{-0.8cm} 
\end{figure*}    
\subsubsection{Performance of Learned Assistance Profile} 
The effectiveness of the learned assistive torque profile was evaluated against five state-of-the-art baseline methods under varying walking speeds: 
\begin{itemize} 
\item \textbf{Baseline \#1 (Spline Controller):} The assistive torque profile was parameterized using spline functions following the implementation in  ~\cite{franks2021comparing}. 
\item \textbf{Baseline \#2 (DOFC):} Delayed output feedback control~(DOFC) was implemented following~\cite{lim2019delayed}, with the control gain $k=13.0$ and time delay $\Delta_{\mathrm{delay}}=0.25$s. 
\item \textbf{Baseline \#3 (BTDC):} A biological torque delayed controller~(BTDC) was implemented to generate the assistive torque profile by scaling and delaying the biological joint moment~\cite{molinaro2024task}, which was obtained from the musculoskeletal simulation. 
\item \textbf{Baseline \#4 (MLP-PPO):} An MLP-based exoskeleton control policy was trained using PPO following the implementation in~\cite{luo2024experiment}. 
\item \textbf{Baseline \#5 (LSTM-PPO):} An LSTM-based exoskeleton control policy was trained using PPO following~\cite{barati2026end}, where an LSTM-based sequence encoder was incorporated to capture temporal information. 
\end{itemize} 
% \end{itemize} 
% \begin{figure}[t]
% \centering
% \setlength{\abovecaptionskip}{-0.04cm} 
% \includegraphics[
%     width=\columnwidth,
%     trim=2mm 1mm 0mm 1mm,
%     clip
% ]{figures/sim_activation_rms_AVG.pdf}
% \caption{Comparison of RMS muscle activation across assistance methods at different walking speeds.} 
% \label{fig:muscle_activation} 
% \vspace{-0.8cm}  
% \end{figure} 

MLP-PPO and LSTM-PPO policies were trained at a fixed walking speed of $1.25~\mathrm{m/s}$ following the authors' implementation. 
%The assistive torque profiles and pelvis walking speeds obtained using all baseline methods and our method without online optimization are shown in Fig.~\ref{fig:exo-performance-variable-speed}(a). 
%The positive mechanical power, which reflects the performance of assistance timing, is plotted in Fig.~\ref{fig:exo-performance-variable-speed}(b). 
%Our method achieves higher positive mechanical power than all baseline methods. 
The assistive torque profiles and pelvis walking speeds are shown
in Fig.~3(a), where our method adapts torque magnitude with
walking speed. Fig.~3(b) shows the corresponding positive
mechanical power; with torque amplitudes comparable to DOFC and
the spline controller, our method yields higher power, suggesting
improved assistance timing.
The mean and standard deviation of the activations of two representative muscles~(\textbf{RF: Rectus Femoris; BF: Biceps Femoris}) over eight gait cycles under three walking speeds are calculated and presented in Fig.~\ref{Fig6:comparison-with-baselines}(b). 
The results show that our method consistently reduces RF activation across the evaluated walking speeds. 
% At a target speed of $0.9$, TCN yielded $-59.78\%$ for the rectus femoris (RF) and $+4.25\%$ for the short head of the biceps femoris (BF). MLP-PPO yielded $-40.51\%$ for RF and $-6.47\%$ for BF, while LSTM-PPO yielded $-37.82\%$ for RF and $-9.96\%$ for BF. 
% At a target speed of $1.25$, TCN yielded $-61.27\%$ for RF and $+0.35\%$ for BF. MLP-PPO yielded $-46.75\%$ for RF and $-6.24\%$ for BF, while LSTM-PPO yielded $-47.34\%$ for RF and $-3.50\%$ for BF. 
% At a target speed of $1.6$, TCN yielded $-57.76\%$ for RF and $-2.46\%$ for BF. MLP-PPO yielded $-44.93\%$ for RF and $-7.51\%$ for BF, while LSTM-PPO yielded $-47.40\%$ for RF and $-5.43\%$ for BF. 
% \input{tables/tab_subjects_experiments} 
% \vspace{-0.2cm}  
\section{Experimental Validation}\label{sec-experimental-validation}  
The experiments consisted of treadmill and free walking trials to evaluate the effectiveness of the learned assistance timing and personalized assistance magnitude, respectively. 
\subsection{Experimental Setups}  
\subsubsection{Hardware System}\label{subsec-exp-setups}   
We developed a hip exoskeleton to provide bilateral assistance~(see Fig.~\ref{fig:overview-framwork}(d)). 
Two wearable IMUs~(Chenyi Technology, China) measure bilateral hip kinematics at 100 Hz. 
The distilled TCN-based exoskeleton control policy was directly run on the Raspberry Pi 5 to generate assistive torque profiles. Online optimization was conducted using the GUI and voice interaction. 
Wireless EMG sensors~(Trigno IM, Delsys Inc., USA) were attached to measure muscle activations. 
\subsubsection{Experimental Protocol}\label{subsec-experimental-protocol} 
Six able-bodied participants (five males and one female) were recruited to evaluate the proposed method, with an average age of 26.3 $\pm$ 3.4 years, body mass of 66.2 $\pm$ 7.4 kg, and height of 175.7 $\pm$ 4.5 cm. The experimental protocol was approved by the Institutional Review Board~(IRB) of XXX. 
All participants were informed of the experimental procedures and potential risks and provided written informed consent. 
The mechanical power was evaluated for all participants under each walking condition. 
In addition, the muscle activations of BF and RF were measured for each participant at three walking speeds, 0.9m/s, 1.25m/s, and 1.6m/s. 
The evaluation metrics include root-mean-square torque~($\btau_{\mathrm{RMS}}$), maximum torque~($\btau_{\mathrm{MAX}}$), mean positive power~(MPP), and mean negative power~(MNP).  
% Due to hardware limitation, our method with online optimization only test the performance of the baseline DOFC~\cite{lim2019delayed} and our method without online optimization. 
% As depicted in Algorithm~\ref{alg:gp_preference}, the online optimization is run $J=30$ iterations and $M=30$ comparison pairs are predefined. 
% Each human subject was asked to select their preferred assistive torque profile from the given options. 
% For each assistive torque profile, the subject walked for 30$s$, followed by 5$s$ free walking. 
% The subject will walk using the second assistive torque profile for another 30$s$ walking, followed by 10$s$ free walking to collect the response. 

Due to hardware limitations, the online-tuning condition was experimentally compared only with DOFC ~\cite{lim2019delayed} and our method without online personalization. 
The procedure for online optimization of the assistance magnitude is summarized in Algorithm~\ref{alg:gp_preference}. The online optimization was performed for $J=$30 iterations using $M=$50 predefined comparison pairs. 
% At each iteration, each participant sequentially experienced two candidate assistive torque profiles~($\bz_{A}$ and $\bz_B$) and selected the preferred one based on the perceived assistance. 
For each comparison, the participant walked with the first assistive torque profile~($\bz_{A}$) for 30$\mathrm{s}$, followed by 5$\mathrm{s}$ of unassisted walking. 
The participant then walked with the second assistive torque profile~($\bz_B$) for another 30$\mathrm{s}$, followed by 10$\mathrm{s}$ of unassisted walking, during which the participant’s preference between the two profiles was collected. 
\subsection{Experiment Results}\label{subsec-human-study} 
\subsubsection{Treadmill Walking} 
The performance of each method was evaluated by treadmill walking. Each human subject wearing our hip exoskeleton was instructed to walk on a treadmill at the speed of 0.9$m/s$, 1.25$m/s$, and $1.6m/s$. 
For each walking speed, all three controllers were randomly tested for 30$s$. 
All metrics were calculated for the left leg over five gait cycles segmented by peak hip flexion angle. 
The results summarized in Table~\ref{tab:experimental_results} show that our method achieves higher positive mechanical power than DOFC. Furthermore, incorporating online personalization further increases positive mechanical power by adapting the assistance magnitude to individual preferences. 
Muscle activation of RF and BF was measured and analyzed. 
As shown in Fig.~\ref{Fig6:-emg-all}, several participants exhibited changes in BF activation that differed from those observed in simulation. These results suggest that increased positive mechanical power does not necessarily translate into reduced muscle activation. 
\subsubsection{Free Walking}  
Four participants were instructed to walk freely for $20$s while continuously varying their walking speed at their own pace. For DOFC, the same parameters as those used in simulation were applied to all participants, whereas the individually preferred assistive torque profile was applied for our method. 
%As shown in Fig.~\ref{Fig6:-power-free-walking}, our method reduced the RMS assistive torque from $3.59 \pm 0.14$~Nm to $3.25 \pm 0.28$~Nm and the mean negative power from $1.37 \pm 0.57$~W to $0.61 \pm 0.21$~W ($\sim55\%$ reduction; paired $t$-test, $p=0.030$). 
%The proportion of resistive assistance near motion transitions was further reduced from $51.6 \pm 12.7$\% to $37.9 \pm 8.3$\%, indicating less undesired assistance during free walking. 
As shown in Fig.~4(b), our method reduced negative mechanical
power by 55\% ($p=0.030$) and the resistance ratio by 27\%, despite only a 9\% reduction in RMS torque.
These disproportionate reductions indicate that the improvement
is not merely due to weaker assistance, but to more favorable
temporal coordination of assistive torque with human motion.   
\vspace{-2.5mm} 
\section{Discussion and Conclusion} 
\vspace{-1.5mm} 
This work presents a speed-adaptive hip exoskeleton control framework that combines sim-to-real reinforcement learning with preference-based online personalization. 
Previous studies~\cite{luo2024experiment,barati2026end} demonstrated sim-to-real generalization by directly deploying learned exoskeleton policies at several discrete walking speeds. 
Park \textit{et al.}~\cite{park2026learning} further employed a two-stage curriculum over a wide range of walking speeds and slopes and distilled a student policy to generate torque commands for onboard deployment. However, the approaches in~\cite{barati2026end,park2026learning} rely on reflex-based musculoskeletal models to generate stable baseline gaits. 
In contrast, our context-conditioned RL framework learns both human motion generation and exoskeleton assistance across continuously varying walking speeds. 
Furthermore, the proposed policy distillation effectively addresses partial observability during onboard deployment and achieves better performance than MLP-PPO and LSTM-PPO baselines adopted in prior sim-to-real approaches~\cite{luo2024experiment,barati2026end}. 

Existing sim-to-real RL methods typically deploy assistive profiles learned in simulation directly on physical exoskeletons~\cite{luo2024experiment,barati2026end}. 
The learned profiles in~\cite{park2026learning} were scaled according to each subject’s body weight. 
Most existing HIL optimization methods for lower-limb exoskeletons have been validated under specific locomotion conditions~\cite{park2025human,zhang2017human}. 
These methods typically rely on gait-cycle-based parameterizations~\cite{ramella2025rapid,zhang2024toward}, although their effectiveness has also been demonstrated across multiple locomotion tasks. 
In contrast to existing preference-based optimization methods~\cite{ramella2025rapid,zhang2024toward}, our method accommodates continuously varying walking speeds without requiring gait-cycle detection. 
% In our method, a speed-adaptive parameterization is first learned in simulation, reducing the subsequent online optimization to a low-dimensional latent space. 
Crucially, the speed-adaptive structure learned in simulation compresses subsequent online personalization into a low-dimensional latent space, allowing subject-specific adaptation without re-optimizing the full assistive profile.
% This work integrates sim-to-real RL with online personalization to reduce the dimensionality of the parameter space for assistive strategy optimization. 
Overall, the proposed framework couples continuous-speed sim-to-real adaptation with low-dimensional preference-based personalization in a unified control pipeline.
Positive mechanical power and muscle activation of two muscle groups were evaluated to assess its superior performance compared to baselines. 
Several limitations remain to be addressed. 
First, our method was validated only for level-ground walking. Second, its effectiveness was demonstrated only on a hip exoskeleton. 
Future work will investigate its generalization to other locomotion modes, such as slope walking and stair climbing, as well as to exoskeletons targeting other lower-limb joints. 
\vspace{-0.4cm}  
\section{Appendix}  
% \input{tables/tab_algorithm}     
% \begin{table}[htbp]
% \centering
% \caption{Reward weights for different policy-learning stages.
% Velocity and acceleration terms are activated by the hold and ramp
% phase indicators, respectively.}
% \label{tab:reward_weights}
% \setlength{\tabcolsep}{4.5pt}
% \renewcommand{\arraystretch}{1.05}
% \begin{tabular}{lcccc}
% \hline
% \textbf{Parameter}
% & \begin{tabular}[c]{@{}c@{}}\textbf{Human}\\\textbf{Phase 1}\end{tabular}
% & \begin{tabular}[c]{@{}c@{}}\textbf{Human}\\\textbf{Phase 2}\end{tabular}
% & \begin{tabular}[c]{@{}c@{}}\textbf{Exo}\\\textbf{Phase 1}\end{tabular}
% & \begin{tabular}[c]{@{}c@{}}\textbf{Exo}\\\textbf{Phase 2}\end{tabular}
% \\
% \hline
% $r_v$          & 1.0 & 0.5  & 0.5  & 3/2 \\
% $r_{\bar v}$   & 0   & 0.5  & 0.5  & 3/2 \\
% $r_{\dot v}$          & 0   & 0.5  & 0.5  & 3.0 \\
% $r_{\overline{\dot v}}$   & 0   & 0.5  & 0.5  & 3.0 \\
% $r_m$          & 0.1 & 0.05 & 0.15 & 0.15 \\
% $r_{\Delta m}$ & 0.1 & 0 & 0.05 & 0.05 \\
% $r_{\mathrm{constraint}}$          & 1.0 & 0  & 0.5  & 0.5 \\
% $r_{\mathrm{foot}}$          & 0.5 & 0  & 0.3  & 0.3 \\
% $r_q$          & 3.9 & 0    & 0    & 0 \\
% $r_{\dot q}$   & 1.0 & 0    & 0    & 0 \\
% $r_P$          & --  & --   & 4.0  & 0 \\
% $r_{\mathrm{hip}}$
%                & --  & --   & 0    & 2.5 \\
% $r_A$          & --  & --   & 0    & 2.0 \\
% $r_{\Delta u}$ & --  & --   & 0    & 0.5 \\
% \hline
% \end{tabular}
% \end{table} 

\begin{table}[htbp]
\centering
\caption{Reward Weights for All different policy-learning stages.}
\label{tab:reward_weights}
\setlength{\tabcolsep}{4.5pt}
\renewcommand{\arraystretch}{1.05}
\begin{tabular}{lcccc}
\toprule
& \multicolumn{2}{c}{Human} & \multicolumn{2}{c}{Exo}\\
\cmidrule(lr){2-3} \cmidrule(lr){4-5}
\textbf{Parameter} & \textit{Phase} 1 & \textit{Phase} 2 & \textit{Phase} 1 & \textit{Phase} 2 \\
\midrule
$r_v$          & 1.0 & 0.5  & 0.5  & 3/2 \\
$r_{\bar v}$   & 0   & 0.5  & 0.5  & 3/2 \\
$r_{\dot v}$ and $r_{\overline{\dot v}}$         & 0   & 0.5  & 0.5  & 3.0 \\
% $r_{\overline{\dot v}}$   & 0   & 0.5  & 0.5  & 3.0 \\
$r_m$          & 0.1 & 0.05 & 0.15 & 0.15 \\
$r_{\Delta m}$ & 0.1 & 0 & 0.05 & 0.05 \\
$r_{\mathrm{constraint}}$          & 1.0 & 0  & 0.5  & 0.5 \\
$r_{\mathrm{foot}}$          & 0.5 & 0  & 0.3  & 0.3 \\
$r_q$          & 3.9 & 0    & 0    & 0 \\
$r_{\dot q}$   & 1.0 & 0    & 0    & 0 \\
$r_P$          & --  & --   & 4.0  & 0 \\
$r_{\mathrm{hip}}$
               & --  & --   & 0    & 2.5 \\
$r_A$          & --  & --   & 0    & 2.0 \\
$r_{\Delta u}$ & --  & --   & 0    & 0.5 \\
\hline
\end{tabular}
\vspace{-3mm} 
\end{table}  
\begin{algorithm}[htbp]   
\caption{Online Preference-based Optimization}
\label{alg:gp_preference} 
\begin{algorithmic}[1] 
% \STATE \textbf{Exoskeleton Control Policy Distillation}   
% \STATE Collect and resample teacher demonstrations at 100 Hz to construct distillation dataset $\mathcal D_{\mathrm{dist}}$
% \STATE Set distillation steps $T_D$
% \FOR{$t=1$ to $T_D$}
%     \STATE Train student policy $\pi_e^{real}$ on $\mathcal D_{\mathrm{dist}}$
% \ENDFOR  
\STATE \textbf{Initialization:} 
\STATE Pre-train exoskeleton control policy $\bpi_e(\cdot;\btheta_e^{*})$ 
\STATE Parameter space $\bz \in \boldsymbol{\mathcal{Z}}$; 
\STATE Kernel function $k(\cdot,\cdot)$, preference noise $\sigma_p$, exploration coefficient $\beta$, maximum number of queries $K$ 
\STATE Preference dataset $\mathcal{D}_{\textbf{Pre}} \leftarrow \varnothing$ 
\STATE Initialize latent utility
$f(\bz) \sim \mathcal{GP}\!\left(0, k(\bz,\bz') \right)$
\FOR{$k = 1,\ldots,K$} 
% \STATE Infer latent utility posterior from pairwise preferences: 
% \STATE \hspace{\algorithmicindent}
% $ 
% p(f \mid \mathcal{D})
% \propto
% p(f)
% \prod_{(\bz_i \succ \bz_j)
% \in \mathcal{D}}
% \Phi\!\left(
% \frac{
% f(\bz_i)
% -
% f(\bz_j)
% }{\sigma_p}
% \right)
% $
\STATE Approximate $p(f\mid\mathcal{D}_{\textbf{Pre}})$ using Laplace approximation
\STATE Obtain $\mu_k(\bz)$ and $\sigma_k(\bz)$
\STATE Select current best parameter:
\STATE \hspace{\algorithmicindent}
$\displaystyle
\bz_k^A
\gets
\arg\max_{\bz \in \boldsymbol{\mathcal{Z}}}
\mu_k(\bz)
$
\STATE Select challenger with exploration: 
\STATE \hspace{\algorithmicindent}
$\displaystyle
\bz_k^B
\gets
\arg\max_{
\bz \in \boldsymbol{\mathcal{Z}}}
\left[
\mu_k(\bz)  
+
\beta\sigma_k(\bz)
\right]
$
\STATE Evaluate assistive profiles generated by $\bz_k^A$ and $\bz_k^B$  
\STATE Collect user's pairwise preference 
$y_k \in \{A,B\}$ 
\IF{$y_k = A$}
\STATE
$\mathcal{D}_{\textbf{Pre}}
\gets
\mathcal{D}_{\textbf{Pre}} 
\cup
\{(\bz_k^A
\succ
\bz_k^B)\}$
\ELSE
\STATE
$\mathcal{D}_{\textbf{Pre}} 
\gets
\mathcal{D}_{\textbf{Pre}} 
\cup
\{(\bz_k^B
\succ
\bz_k^A)\}$
\ENDIF 
\ENDFOR
\STATE
Obtain best personalized parameter : 
\STATE
$
\bz^{*}
\gets
\arg\max_{\bz\in\boldsymbol{\mathcal{Z}}} 
\mu_K(\bz)
$
\end{algorithmic}
% \vspace{-3mm} 
\end{algorithm}    
\begin{table}[htbp]
\centering
\caption{Hyperparameters of PPO used for all policies learning.} 
\label{tab:ppo_parameters}
\begin{minipage}{0.94\columnwidth}
\centering
\footnotesize
\setlength{\tabcolsep}{1.5pt}
\renewcommand{\arraystretch}{1.05} 
\begin{tabular}{@{}lcccc@{}}
\toprule
& \multicolumn{2}{c}{Human} & \multicolumn{2}{c}{Exo}\\
\cmidrule(lr){2-3} \cmidrule(lr){4-5}
\textbf{Parameter} & \textit{Phase} 1 & \textit{Phase} 2 & \textit{Phase} 1 & \textit{Phase} 2 \\
\midrule
Learning rate
& $1\times10^{-4}$
& $1\times10^{-4}$
& $3\times10^{-5}$
& $3\times10^{-5}$ \\

$n_{\mathrm{steps}}$
& 512 & 512 & 2048 & 2048 \\

Batch size
& 8192 & 8192 & 8192 & 16384 \\

$n_{\mathrm{epochs}}$
& 30 & 30 & 20 & 20 \\

$\gamma$
& 0.99 & 0.99 & 0.99 & 0.99 \\

$\lambda_{\mathrm{GAE}}$
& 0.95 & 0.95 & 0.95 & 0.95 \\

Clip range $\epsilon$
& 0.20 & 0.20 & 0.15 & 0.15 \\

Entropy coefficient
& 0.001 & 0.001 & 0.003 & 0.003 \\

Value function coefficient
& 0.5 & 0.5 & 0.5 & 0.5 \\

Maximum gradient norm
& 0.5 & 0.5 & 0.5 & 0.5 \\

Target KL divergence
& 0.01 & 0.01 & 0.01 & 0.01 \\
\hline
\end{tabular}
% \vspace{3pt} 
\end{minipage}
\end{table}

\vspace{-5mm}  
\bibliographystyle{IEEEtran}   
\bibliography{
IEEEabrv,
utils/references_exo,
utils/my_references
}   
\end{document}